\documentclass[runningheads]{llncs}

\usepackage{graphicx}
\usepackage{url}
\usepackage{color}
\usepackage{hyperref}
\usepackage[table,xcdraw]{xcolor}
\usepackage{amsmath}

\usepackage{amsfonts}
\usepackage{caption}
\usepackage{graphicx}
\usepackage{booktabs}
\usepackage{bbm}
\usepackage{fontawesome5}
\usepackage{wrapfig}
\usepackage{multirow}
\usepackage{capt-of}
\definecolor{olive}{rgb}{0.6, 0.6, 0.2}
\definecolor{sand}{rgb}{0.8666666666666667, 0.8, 0.4666666666666667}
\definecolor{wine}{rgb}{0.5333333333333333, 0.13333333333333333, 0.3333333333333333}
\definecolor{deblue}{RGB}{11,132,147}
\definecolor{ocra}{RGB}{204, 119, 34}
\usepackage{enumitem}
\usepackage{tikz}
\newcommand{\fcircle}[2][red,fill=red]{\tikz[baseline=-0.5ex]\draw[#1,radius=#2] (0,0.03) circle ;}

\begin{document}
\title{Bilevel Hypergraph Networks for Multi-Modal Alzheimer's Diagnosis }

\author
{%Anonymous Submission MICCAI24
Angelica I. Aviles-Rivero\inst{1} \and
Chun-Wun Cheng  \inst{1} \and
Zhongying Deng \inst{1}   \and \\
Zoe Kourtzi  \inst{2}   \and
Carola-Bibiane Schönlieb\inst{1}\thanks{the Alzheimer’s Disease Neuroimaging Initiative}
}
\authorrunning{Aviles-Rivero et al.}
\institute{
DAMTP, University of Cambridge, UK  \\
\email{\{ai323,cwc56,zd294,zk240,cbs31\}@cam.ac.uk}\and
Department of Psychology, University of Cambridge, UK \{\email{zk240@cam.ac.uk}\}
}
\maketitle              % typeset the header of the contribution
\begin{abstract}
Early detection of Alzheimer's disease's precursor stages  is imperative for significantly enhancing patient outcomes and quality of life. This challenge is tackled through a semi-supervised multi-modal diagnosis framework. In particular, we introduce a new hypergraph framework that enables higher-order relations between multi-modal data, while utilising minimal labels. We first introduce a bilevel hypergraph optimisation framework that jointly learns a graph augmentation policy and a semi-supervised classifier.  This dual learning strategy is hypothesised to enhance the robustness and generalisation capabilities of the model by fostering new pathways for information propagation.  Secondly, we introduce a novel strategy for generating pseudo-labels more effectively via a gradient-driven flow. Our experimental results demonstrate the superior performance of our framework over current techniques in diagnosing Alzheimer's disease.

\keywords{Hypergraph Learning \and Multi-modal Analysis \and Semi-supervised Learning \and Alzhei\-mer's disease \and Bilevel Optimisation}
\end{abstract}
\section{Introduction}
Alzheimer's disease, a neurodegenerative condition, remains incurable. However, the silver lining in this challenging scenario is the potential for early detection, which can significantly enhance the quality of life for patients by enabling timely treatment interventions. The quest for automated diagnosis of Alzheimer's and its prodromal stages has seen considerable exploration within the literature~\cite{zhou2021synthesizing,shin2020gandalf,aviles2022multi,liang2023modeling,wang2023hgib}. 
While existing research highlights the promise of deep learning techniques in addressing this challenge, two significant barriers hinder progress. First, the integration of diverse data modalities holds the potential to improve  diagnosing AD. Yet, current methods fall short in harnessing the full spectrum of this data, primarily due to the complexity of forging meaningful connections across the multimodal landscape. Second, there's a pressing demand for models that demand minimal supervision, navigating around the constraints of time, cost, and potential bias inherent in data labelling. This paradigm shift towards leveraging minimal supervision could dramatically streamline the path towards more accurate diagnostic for Alzheimer's research.

A body of literature has explored hypergraph learning~\cite{pan2021characterization,zuo2021multimodal,shao2020hypergraph,shao2021hyper,aviles2022multi} to overcome the aforementioned challenges. Hypergraphs are a generalisation of graphs where an edge can connect a set of nodes, modelling beyond pair-wise relations. Owing to this property, hypergraphs can capture higher-order relations across multi-modal data. However, the limitations of the body of literature are two-fold. Firstly, existing works are based on similar principles following the functional of that~\cite{zhou2006learning} -- even deep hypergraph frameworks,  e.g.,~\cite{yadati2019hypergcn,feng2019hypergraph}. Secondly, there is a scarcity of works to explore the influence of topological and feature augmentation on hypergraph learning.  Existing techniques are mainly based on data augmentation~\cite{cubuk2020randaugment}, but graphs/hypergraphs cannot be augmented similarly as they need certain topological and feature augmenters. Whilst existing literature covers graph augmentation extensively~\cite{you2021graph,zhao2022autogda,ding2022data}, these methodologies are primarily tailored for graphs and do not fully address the complexities inherent in hypergraphs. The field lacks in-depth exploration of hypergraph-specific augmentation methods, with few existing studies often requiring manual intervention~\cite{aviles2022multi}.

\textbf{Contributions.} In this work, we introduce a novel semi-supervised hypergraph framework for multi-modal analysis, in which we highlight two major contributions. \fcircle[fill=wine]{2.5pt} First, we propose a bilevel hypergraph optimisation framework that jointly learns  a graph augmentation policy and a semi-supervised classifier. Our hypothesis is that hypergraph augmentations can forge new pathways for information propagation yielding to higher robustness and generalisation. Particularly, as hypergraphs exploit higher-order relations such interactions can lead to resilience against errors. \fcircle[fill=wine]{2.5pt} Second, we introduce a strategy for pseudo-labels via a gradient-driven flow. We show that our bilevel frameworks demonstrate superior performance than existing techniques for Alzheimer's disease diagnosis.

\begin{figure}[t!]
\centering
\includegraphics[width=1\textwidth]{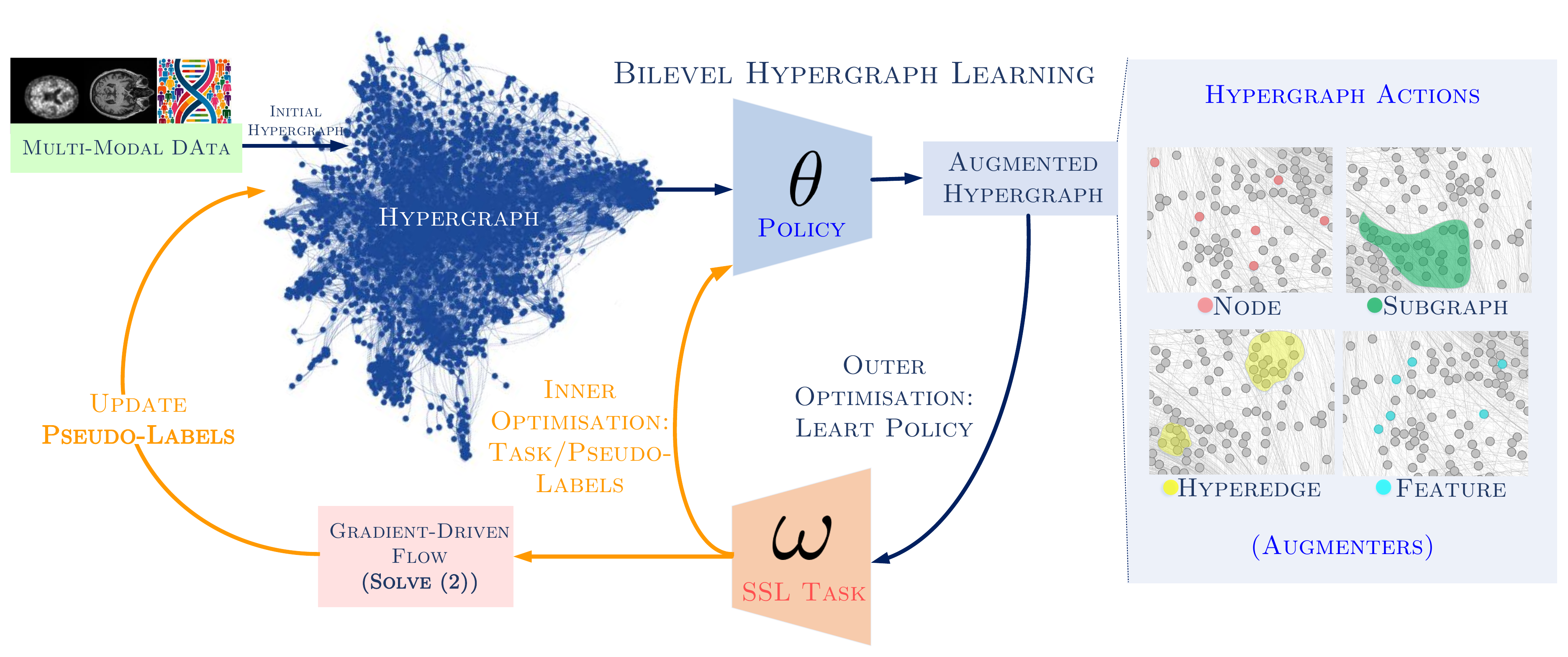}
\caption{Our Bilevel Hypergraph Learning Framework for Early Alzheimer's Disease Diagnosis. It integrates multi-modal data into a hypergraph, employing a bilevel optimisation strategy to co-learn a graph augmentation policy and a semi-supervised classifier. It introduces an innovative pseudo-label update mechanism via gradient-driven flow, aiming for enhanced learning with minimal labels. }
\label{fig::teaser}%\vspace{-0.9cm}
\end{figure}

\section{Methodology}
This section introduces our framework for learning through semi-supervised hypergraphs, which overview is displayed in Figure~\ref{fig::teaser}.

%\smallskip
\subsection{Multi-Modal Hypergraph Construction \& Actions}
We address the challenging problem of multi-modal analysis for Alzheimer's diagnosis by considering an ensemble of $\mathbf{M}$ different data modalities. For each modality, our dataset comprises $N$ individual observations, denoted as $X = \{x_1,...,x_N\} \in \mathcal{X}$, each of which is sampled from a  probability distribution $\mathbb{P}$ across the domain $\mathcal{X}$. This approach yields a composite dataset structured as ${X_1,...,X_{s},X_{s+1},..., X_\mathbf{M}}$, where the dataset is composed of two main types: the first $s$ modalities include various forms of imaging data, such as MRI scans, which are critical for visual assessment of the brain, and the subsequent modalities encompass non-imaging data, like demographics information.

\fcircle[fill=wine]{2.5pt} \textbf{From Multi-Modal Data to Hypergraphs.} Consider a hypergraph defined as $\mathcal{G}=(\mathcal{V},\mathcal{E},\mathcal{W})$, where $\mathcal{V} =\{v_1,...,v_n\} $, with $|\mathcal{V}|= n$  denoting the set of vertices, and $\mathcal{E}= \{e_1, ..., e_m\}$, with $|\mathcal{E}| = m$ representing the hyperedges. Additionally, $\mathcal{W} : \mathcal{E}\to\mathbb{R}_{>0}$ is a function assigning positive real weights to each hyperedge $e$, which encompasses a subset of vertices. The corresponding incidence matrix is denoted as
$\mathcal{H}_{a,e} = \begin{cases} 
1 & \text{if } a \in e, \\
0 & \text{otherwise}
\end{cases}
$
, for $a \in \mathcal{V}$ and $e$ indicating a hyperedge, essentially a collection of vertices from $\mathcal{V}$. 

We start by performing feature extraction on the imaging data. A transformation $f_{\omega}$ is applied to map $X$ into a vector space denoted by $\mathbf{v} = \{\mathbf{v}_1, \ldots, \mathbf{v}_N\}$ with each element $\mathbf{v}_i = f_{\omega}(x_i)$.
The k-nearest neighbors of a given embedding $\mathbf{v}_i$ are represented as $\text{NN}_k(\mathbf{v}i)$. For each dataset $X_1, \ldots, X_s$, we construct a unique hypergraph $\mathcal{H}_{ij}^{1,\ldots,s}$, defined by the rule: $\mathcal{H}_{ij}^{1,\ldots,s} = [\mathbf{v}_i^\top \mathbf{v}_j]$ if $\mathbf{v}_i$ belongs to $\text{NN}_k(\mathbf{v}_j)$; otherwise, it is set to 0. Similarly, for non-imaging data, we evaluate the similarity between subjects $\mathbf{x}$ and their respective phenotypic metrics $\mathbf{z}$ to produce $\mathcal{H}^{s+1,\ldots,\mathbf{M}}$. This involves calculating $S(\mathbf{x},\mathbf{z})$, where $S$ is a similarity function. Here, $S(\mathbf{x},\mathbf{z})$ is determined if $\mathbf{x}$ falls within the $k$-nearest neighbors of $\mathbf{z}$; if not, the value is set to 0. The construction of the overarching hypergraph $\mathcal{H}$ is accomplished through the concatenation of hypergraph representations from each modality, denoted $\mathcal{H}^1$ through $\mathcal{H}^{\mathbf{M}}$, effectively encompassing the spectrum of imaging and non-imaging data modalities.

\fcircle[fill=wine]{2.5pt} \textbf{Hypegraph Actions.} After defining the construction of our multi-modal hypergraph, and before we proceed to introduce the proposed hypergraph semi-supervised bilevel optimisation scheme. A key element in our framework is a learned policy (details in the following section), which hinges on learned topological and feature augmentations within the hypergraph. While graph augmentation policies have been explored previously, research into hypergraph topology and features—and their benefits for complex tasks such as multi-modal diagnosis—remains sparse. \textit{Our hypothesis posits that hypergraph augmentation, unlike simple data augmentation, can forge new pathways for information propagation. This facilitates the learning of more robust representations that are not strictly bound by the original data distribution. Moreover, the exploitation of higher-order interactions among different data modes (e.g., MRI images, PET scans, genetic information, cognitive tests) leads to resilience against errors or inconsistencies in individual data modalities.} 

We define a specific set of manipulations on graphs through a function $\psi: \mathcal{X} \times \mathcal{G} \rightarrow \mathcal{G}$. This function is selected from the set  $\psi \in \{\alpha_\mathcal{V}, \rho_\mathcal{V}, \rho_{\mathcal{E}\mathcal{V}}, \rho_\mathcal{E}\}$, with each element representing a distinct type of manipulation based on parameters $X \in \mathcal{X}$. 
\renewcommand{\labelenumi}{\fcircle[fill=deblue]{2.5pt}}
        \begin{enumerate}
            \item {Node Removal ($\alpha_\mathcal{V}$):} remove nodes in a set $\mathcal{X} $ such that $ \mathcal{X} =\mathcal{R}_\mathcal{V} \in \mathcal{V}$, $\alpha_\mathcal{V}(\mathcal{R}_\mathcal{V}, \mathcal{G}) = (\mathcal{V} \backslash \mathcal{R}_\mathcal{V}, \mathcal{E})$ 
            \item  {Hyperedge Removal ($\rho_\mathcal{E}$):} remove hyperedges in a set $\mathcal{X}$ such that $ \mathcal{X} = \mathcal{R}_\mathcal{E} \in \mathcal{E}$,  $\rho_\mathcal{E}(\mathcal{R}_\mathcal{E}, \mathcal{G}) = (\mathcal{V}, \mathcal{E} \backslash \mathcal{R}_\mathcal{E})$. 
            \item  {Subgraph Removal ($\rho_{\mathcal{E}\mathcal{V}}$): }hyperedges subgraph $\mathcal{X}$ such that $\mathcal{X} = \{\mathcal{R}_{\mathcal{E}} \in \mathcal{E}, \mathcal{R}_{\mathcal{V}} \in \mathcal{V}\}$, $\rho_{\mathcal{E},\mathcal{V}}(\mathcal{R}_\mathcal{E},\mathcal{R}_\mathcal{V} \mathcal{G}) = (\mathcal{V}  \backslash  \mathcal{R}_\mathcal{V},  \mathcal{E} \backslash \mathcal{R}_\mathcal{E})$ 
            \item { Feature Perturbation ($\rho_\mathcal{V}$): } $\delta-$Feature Perturbation in a set $\mathcal{X} $ such that $\mathcal{X} = \mathcal{P}_\mathcal{V}$, $\alpha_\mathcal{E}(\mathcal{P}_\mathcal{V}, \mathcal{G}) $
        \end{enumerate}
In this work, we introduce two types of actions. Firstly, topological augmentation includes node, hyperedge and subgraph removal. Secondly, we also consider feature perturbation, which offers a different type of possible augmentation.  The set  $\psi$ is the searched ratio for each action. For clarity purposes, the subgraph action is given by a random walk determined by $\rho_{\mathcal{E},\mathcal{V}}$.

\subsection{Bilevel Hypergraph Optimisation}
We now detail our augmented semi-supervised hypergraph framework, designed to tackle the complexities of Alzheimer's disease diagnosis. At the heart of our framework lies a bilevel optimisation scheme, crafted to learn a semi-supervised classifier for diagnosis alongside a graph augmentation policy. This dual learning objective sets our approach apart from existing techniques. Our designed framework not only enhances the hypergraph's ability to generalise by introducing new propagation pathways but also facilitates deeper exploitation of higher-order relationships within the multi-modal data.

Our framework utilises a set of labelled data, \(X_l = \{ (x_i ,y_i) \}_{i=1}^{l}\), where \(y_i\) is the label of data $x_i$, alongside a vast set of unlabeled data, \(X_u = \{ x_k \}_{k=l+1}^{m}\). We then propose the following optimisation scheme, which reads:

\faArrowCircleRight  \textbf{ Inner Optimisation  (Semi-Supervised Classifier with Gradient -Based-Flow Pseudo-Labels).} Our semi-supervised hypergraph framework is based on the principle of pseudo-labelling. This lower-level optimisation aims to address our main downstream task: classification (diagnosis). Accordingly, the loss function is defined as:
\begin{equation} \label{gradient-based-flow}
    \omega^* = \arg\min_{\omega} \mathcal{L}_{lab}(f_{\omega}(A_{\theta}(X_l)), Y_l) +  \tau \mathcal{L}_{unc}(f_{\omega}(A_{\theta}(X_u)), \hat{Y}_u),
\end{equation}
where $\mathcal{L}_{lab}$ and $\mathcal{L}_{unc}$ are the cross-entropy loss for the labelled and unlabelled data respectively. $\tau$ is the measure of uncertainty via entropy ~\cite{kendall2017uncertainties,abdar2021review} such that $\tau + \frac{Q(\hat{Y}_u)}{log(l)} = 1$, where Q is the entropy and $\hat{Y}_u$ is normalised beforehand. $f_{\omega}$ is the classification networks parameterized by $\omega$. \( A_{\theta} \) denotes the augmentation function parameterised by \( \theta \), \( X_l \) and \( Y_l \) are the labelled data and their corresponding labels, \( X_u \) is the unlabelled data, and \( \hat{Y}_u \) are the generated pseudo-labels. \textbf{How is $\hat{Y}_u$ inferred?} Unlike existing semi-supervised pseudo-labelling models, our approach infers the pseudo labels via a gradient-based flow. We aim to minimise the total variation function $TV_{J}(u_k)$ using the following  semi-explicit flow, and we update the numerical solution $u_{k}$ by iteration k :
\begin{equation}
\begin{cases} 
u_{k+\frac{1}{2}} \|u_{k}\|  = \|u_{k}\|u_k + \Delta t \left( TV_{J}(u_k) (c_k - \tilde{c}_k) - \|u_{k}\|\gamma _{k+\frac{1}{2}} \right)  \\
u_{k+\frac{1}{2}} = u_{k+1}\|u_{k+\frac{1}{2}}\|_2 
\end{cases}
\label{eq:pde}
\end{equation}
where $\gamma _{k+\frac{1}{2}} \in \partial TV_{J}(u_{k+\frac{1}{2}})$, $c \in \partial \lVert u^k \rVert$. Here, $\partial f$ represents the set of potential subdifferentials of a convex function $f$, defined as follows: $\partial f = \{ \gamma  \mid \exists u, \text{with } \gamma  \in \partial f(u) \}$. Moreover, $\Delta t$ is the time step size in the numerical method, which is a positive number since $\Delta t >  0$. The scaling function, denoted as $d(x)$, is utilised to define the scaled quantity $\tilde{c}_k = \frac{\langle d, c_k \rangle}{\langle d, d \rangle} d$. After convergence of \eqref{eq:pde} at \( u^* = [u^{*,1}, \ldots, u^{*,L}] \), we determine the label for each node to be \( \hat{y}_i = \text{argmax}_j \ u^{*,j}_i \). The pseudo-labels generated, represented as \( \hat{Y}_u \), are set to \( \{\hat{y}_k\}_{k=l+1}^{n} \).

\faArrowCircleRight \textbf{ Outer Optimisation (Learnt Augmentation Policy).}
We define an augmentation policy network \( g_{\theta} \) that learns to apply transformations to the hypergraph, following the actions outlined in Section 2.1, to improve the model's performance on hypergraph semi-supervised tasks. Our goal is to optimise the augmentation policy \( \theta \) based on the task's performance. Denote $X=X_l \cup X_u$ and $Y=Y_l \cup \hat{Y}_u$, the outer optimisation is formulated as

\begin{equation}
    \theta^* = \arg\max_{\theta} \mathcal{M_P}(f_{\omega^*}(A_{\theta}(X)), Y),       
\end{equation}
where \( \mathcal{M_P} \) is the performance metric, specifically accuracy.  The optimal model parameters from the inner optimisation are denoted by \( \omega^* \).

\section{Experimental Results}

\textbf{Dataset Description.} We validate our bilevel semi-supervised hypergraph framework to analyse the Alzheimer's Disease Neuroimaging Initiative (ADNI) dataset~\footnote{adni.loni.usc.edu}. This dataset, sourced from various centers, encompasses a rich variety of multi-modal data, including images and diverse phenotype information. Our analysis focuses on a subset of 500 individuals, considering 
Magnetic Resonance Imaging (MRI), Positron Emission Tomography (PET), demographic details, and Apolipoprotein E (APOE). 
We consider four distinct groups (classes): early and late stages of mild cognitive impairment (EMCI and LMCI), normal control (NC), and Alzheimer's disease (AD).

\textbf{Comparative Analysis and Metrics.} Our methodology for evaluating our framework involves a direct comparison with leading hypergraph learning techniques such as HGSCCA~\cite{shao2021hyper}, HGNN~\cite{feng2019hypergraph}, DHGNN~\cite{jiang2019dynamic},  HGNN+~\cite{gao2022hgnn+} and  Dual HG~\cite{aviles2022multi}, alongside a comparison with GNNs~\cite{parisot2018disease}. 
For evaluating the performance, we adhere to the medical field's standard metrics: we measure the Error Rate (ER) along with its average (Avg-ER), and the Positive Predictive Value (PPV), which serves as an equilibrium measure between sensitivity and specificity.

\begin{table*}[t!]
\resizebox{0.98\textwidth}{!}{%
\begin{tabular}{lccccccc}
\hline
\multicolumn{1}{c}{} & \multicolumn{4}{c}{\cellcolor[HTML]{EFEFEF}\textsc{Error Rate} $\downarrow$} &  & \cellcolor[HTML]{EFEFEF} & \cellcolor[HTML]{EFEFEF} \\ \cline{2-5}
\multicolumn{1}{c}{\multirow{-2}{*}{}} & NC & EMCI & LMCI & AD &  & \multirow{-2}{*}{\cellcolor[HTML]{EFEFEF}Avg-ER} & \multirow{-2}{*}{\cellcolor[HTML]{EFEFEF}PPV$\uparrow$} \\ \hline
GNNs~\cite{parisot2018disease} & 27.99$\pm$1.41 & 27.46$\pm$1.21 & 29.31$\pm$1.66 & 26.16$\pm$1.63 &  & 27.73 & 72.37 \\
HGSCCA~\cite{shao2021hyper} & 25.21$\pm$1.20 & 26.53$\pm$1.35 & 27.15$\pm$1.35 & 25.67$\pm$1.48 &  & 26.14 & 73.96 \\
HGNN~\cite{feng2019hypergraph} & 23.10$\pm$1.21 & 24.32$\pm$1.10 & 25.91$\pm$1.18 & 24.01$\pm$1.03 &  & 24.33 & 75.76 \\
DHGNN~\cite{jiang2019dynamic} & 20.52$\pm$0.98 & 22.62$\pm$1.15 & 23.06$\pm$1.10 & 21.16$\pm$0.95 &  & 21.84 & 78.26 \\
HGNN+~\cite{gao2022hgnn+} & 18.41$\pm$1.01 & 18.99$\pm$1.11 & 20.65$\pm$1.54 & 18.99$\pm$0.79 &  & 19.26 & 80.84 \\
Dual HG~\cite{aviles2022multi} & 15.21$\pm$0.64 & 17.52$\pm$0.63 & 18.91$\pm$0.92 & 16.17$\pm$0.56 &  & 16.95 & 83.15 \\
Ours & \cellcolor[HTML]{FFFFD4}12.79$\pm$0.28 & \cellcolor[HTML]{FFFFD4}13.06$\pm$0.34 & \cellcolor[HTML]{FFFFD4}14.66$\pm$0.55 & \cellcolor[HTML]{FFFFD4}12.92$\pm$0.41 &  & \cellcolor[HTML]{FFFFD4}13.36 & \cellcolor[HTML]{FFFFD4}86.74 \\ \hline
\end{tabular}
}
\caption{Numerical comparison of our method against established (graph) and hypergraph techniques, demonstrating the  performance across various metrics. The top-performing results are emphasised with \colorbox[HTML]{FFFFD4}{yellow} highlighting.} \label{table1}
\end{table*}
\begin{table}[t!]
\resizebox{\textwidth}{!}{%
\begin{tabular}{llcccccc}
\hline
 &  & \multicolumn{4}{c}{\cellcolor[HTML]{EFEFEF}\textsc{Error Rate} $\downarrow$} & \cellcolor[HTML]{EFEFEF} & \cellcolor[HTML]{EFEFEF} \\ \cline{3-6}
 & \multirow{-2}{*}{Aug.} & NC & EMCI & LMCI & AD & \multirow{-2}{*}{\cellcolor[HTML]{EFEFEF}Avg-ER} & \multirow{-2}{*}{\cellcolor[HTML]{EFEFEF}PPV$\uparrow$} \\ \hline
Dual HG~\cite{aviles2022multi} & D+HA & 15.21$\pm$0.64 & 17.52$\pm$0.63 & 18.91$\pm$ 0.92 & 16.17$\pm$0.56 &  16.95 & 83.15  \\
 & A0 & 14.03$\pm$0.60 & 15.13$\pm$0.58 & 16.07$\pm$0.66 & 14.16$\pm$0.52 & 14.85 & 85.25 \\
 & A1 & \cellcolor[HTML]{FFE7E6}13.45$\pm$0.34 & \cellcolor[HTML]{FFE7E6}14.06$\pm$0.41 & \cellcolor[HTML]{FFE7E6}15.66$\pm$0.56 & \cellcolor[HTML]{FFE7E6}13.79$\pm$0.45 & \cellcolor[HTML]{FFE7E6}14.24   & \cellcolor[HTML]{FFE7E6}85.86  \\
 & A2 & 13.99$\pm$0.36 & 14.48$\pm$0.47 & 15.97$\pm$0.61 & 14.02$\pm$0.60 & 14.61 & 85.48 \\
 & A3 & 15.16$\pm$0.61 & 16.48$\pm$0.62 & 17.80$\pm$0.71 & 15.76$\pm$0.67 & 16.30 & 83.80  \\
\multirow{-5}{*}{\begin{tabular}[c]{@{}l@{}}Our\\ Policy\end{tabular}} & A4 & \cellcolor[HTML]{FFFFD4}12.79$\pm$0.28 & \cellcolor[HTML]{FFFFD4}13.06$\pm$0.34 & \cellcolor[HTML]{FFFFD4}14.66$\pm$0.55 & \cellcolor[HTML]{FFFFD4}12.92$\pm$0.41 & \cellcolor[HTML]{FFFFD4}13.36  & \cellcolor[HTML]{FFFFD4}86.74 \\ \hline
\end{tabular}
}
\caption{Performance comparison on the influence of various augmentation types. The augmentations include Data Augmentation + Heuristic Hypergraph Augmentation (D+HA), Node Removal (A0), Hyperedge Removal (A1), Subgraph Removal (A2), Feature Perturbation (A3), and an All-in-One Policy (A4). 
%The performance is reported in terms of Error Rate  along with the overall performance (Avg-ER and PPV). 
The best performance is highlighted in \colorbox[HTML]{FFFFD4}{yellow},
and the second-best in \colorbox[HTML]{FFE7E6}{orange}.
}\label{table2} \vspace{-0.2cm}
\end{table}

For the initialisation of the graph construction process, we employed the ResNet-50 architecture as future extraction. We construct our graph by configuring the k-Nearest Neighbors (k-NN) parameter to a value of 25, implemented a weight decay rate of 0.0002, and established an initial learning rate of 0.05, which we gradually reduced according to a cosine annealing schedule over a course of 150 epochs. Consistent with conventional approaches in semi-supervised learning, we conducted five independent selections of labelled samples at random intervals. The results are then presented as the average values of the performance metrics, accompanied by their respective standard deviations.

\fcircle[fill=deblue]{2.5pt} \textbf{Comparison against Existing Techniques.} The results presented in Table~\ref{table1} offer a comprehensive numerical comparison between our proposed method and several established techniques, both from graph-based and hypergraph-based paradigms, across various diagnostic classes. Notably, our method achieves the lowest error rates across all classes, demonstrating superior performance with an Avg-ER of 13.36 and a PPV of 86.74. This indicates a significant improvement in diagnostic accuracy and predictive capability compared to the existing methods.

\begin{figure}[t!]
\centering
\includegraphics[width=1\textwidth]{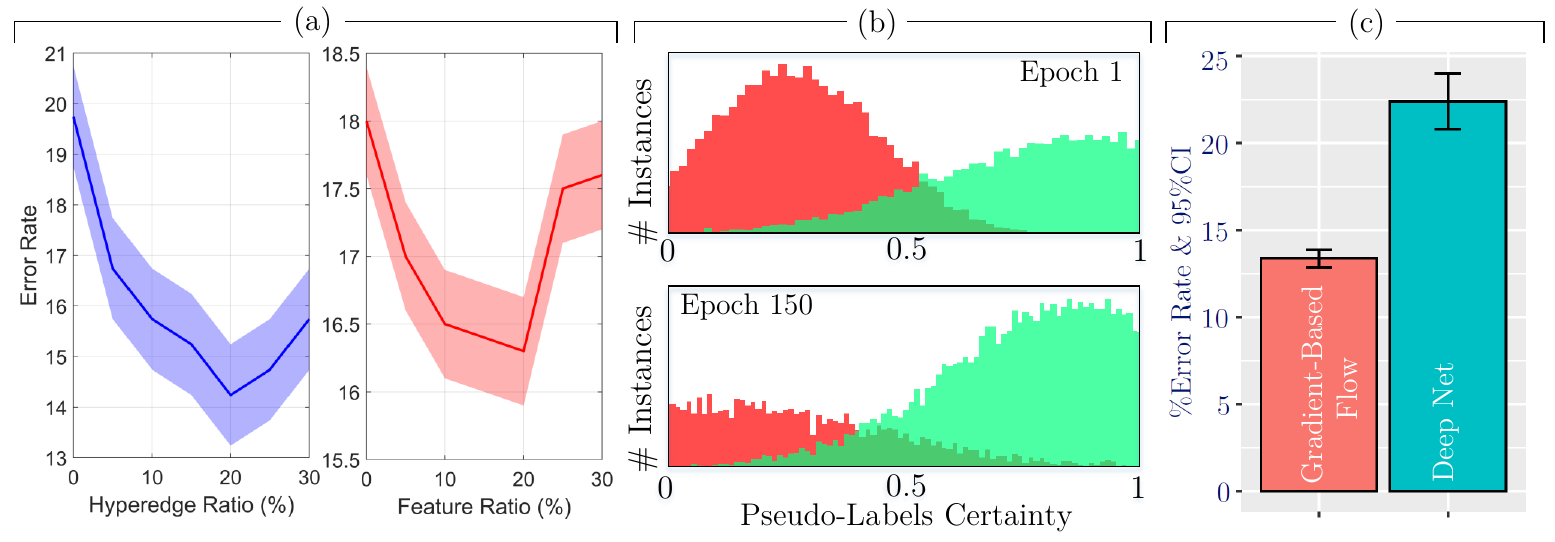}
\caption{\textbf{Ablation Studies.} (a) Demonstrates the impact of learned augmentations within our framework.
(b) Highlights the progression of pseudo-label certainty from Epoch 1 to Epoch 150,
(c)Error rates obtained using our pseudo-labels versus those generated directly from a deep network.
} %\vspace{-0.5cm}
\label{fig::ablation1}
\end{figure}

\begin{figure}[t!]
\centering
\includegraphics[width=0.85\textwidth]{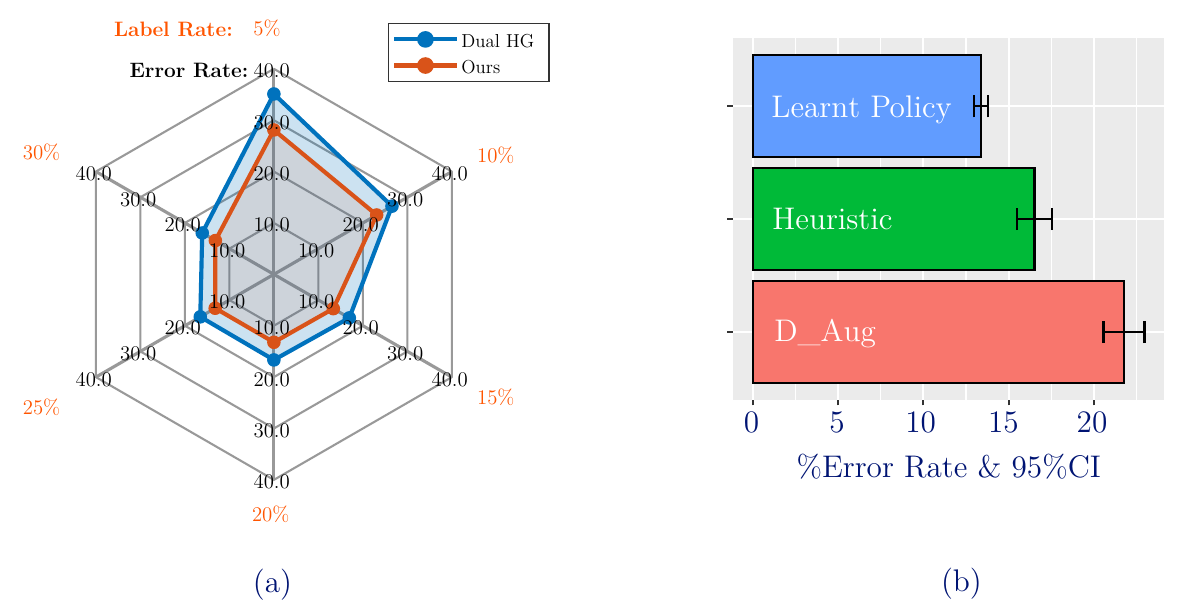}
\caption{(a) Comparison of our technique with Dual HG across label rates, showing error rate differences. (b) Error rate comparison for Learned Policy, Heuristic, and D\_Aug strategies.} \vspace{-0.2cm}
\label{fig::ablation2}
\end{figure}

The comparison reveals that graph neural networks (GNNs)~\cite{parisot2018disease} 
show higher error rates across the board, emphasising the challenges in capturing complex relationships within data using traditional graph structures. On the other hand, hypergraph-based methods 
progressively refine these models by incorporating more nuanced data relationships.
However, these methods still fall short when compared to our approach, underlining the effectiveness of our method in leveraging the inherent data complexity for more accurate disease diagnosis.

\fcircle[fill=deblue]{2.5pt} \textbf{The Role of Learned Augmentations.} 
From Table~\ref{table2}, we observe that Dual HG (second best) utilises both data augmentation and heuristic hypergraph augmentation (D+HA) to enhance model performance. In contrast, our augmentation techniques, integrated within our bilevel framework, offer a novel approach to learning augmentation strategies. Developed within this framework, our proposed augmentations introduce a series of learned augmentation policies (A0 through A4) that dynamically adapt to the data and hypergraph structure. The adaptability and specificity of these augmentations facilitate targeted improvements in error rates and overall performance, as evidenced by the results.
The dynamic adaptability of our strategies, particularly evident in policies like A0 (Node Removal), A1 (Hyperedge Removal), A2 (Subgraph Removal), and A4 (All-in-One Policy, encompassing all augmentations A0-A3), highlights how specific alterations to the hypergraph's structure can lead to substantial improvements in error rates and overall performance metrics, in contrast to the effects of solely relying on  heuristic augmentation (Dual HG).

\fcircle[fill=deblue]{2.5pt} \textbf{Ablation on Pseudo-labels and Learnt Policy.} 
The exploration of learned augmentations (see Figure~\ref{fig::ablation1}-(a)), specifically the Hyperedge Ratio and Feature Ratio, underscores the adaptability and optimisation capability of our framework. The figure showcases the variation in error rates as these ratios are adjusted, highlighting the efficiency of a learned policy over heuristic settings. 
In Figure~\ref{fig::ablation1}-(b), the transition from red to green areas represents the shift from incorrect to correct pseudo-labels relative to the ground truth.
This progression, depicted from Epoch 1 to Epoch 150, validates our contribution towards enhancing the certainty of pseudo-labels over time. The increase in green area over successive epochs reflects the efficacy of our proposed gradient-based flow pseudo-labels (see~\eqref{gradient-based-flow}), marking a significant departure from conventional semi-supervised techniques that typically generate pseudo-labels directly from a deep network. 
These results are further supported in Figure~\ref{fig::ablation1}-(c), where the error rates between our pseudo-labeling technique and those generated directly from a deep network further emphasises our approach.

\fcircle[fill=deblue]{2.5pt} \textbf{Ablation Label Rates and Type of Augmenters.} In Figure~\ref{fig::ablation2}-(a), our technique demonstrates superior performance across varying label rates compared to the Dual HG approach.
Figure~\ref{fig::ablation2}-(b)
validates our hypothesis that learning policies at the hypergraph level can significantly outperform traditional data augmentation methods. By introducing topological augmentations, our approach creates novel pathways for information propagation within the hypergraph. This leads to substantial improvements in error rates.

\section{Conclusion}
Our  work on semi-supervised hypergraph learning for multi-modal data showcases the strength of our novel bilevel optimisation framework and gradient-driven flow. We have demonstrated  superiority over existing methodologies, highlighting the significance of high-order topological augmentations in effectively handling complex data structures. This confirms our assertion that strategic, structure-focused enhancements are crucial for advancing the state of the art in multi-modal hypergraph learning.

%%%%%%%%%%%%%
\section*{Acknowledgments}
AAR gratefully
acknowledges funding from the Cambridge Centre for
Data-Driven Discovery and Accelerate Programme
for Scientific Discovery, made possible by a donation
from Schmidt Futures, ESPRC Digital Core Capability
Award, and CMIH, CCMI, University of Cambridge.
CWC gratefully
acknowledges funding from CCMI, University of Cambridge.
ZD and ZK acknowledges the funding from Wellcome Trust 221633/Z/20/Z. 
CBS acknowledges support
from the Philip Leverhulme Prize, the Royal Society
Wolfson Fellowship, the EPSRC advanced career fellowship EP/V029428/1, EPSRC grants EP/S026045/1 and EP/T003553/1, EP/N014588/1, EP/T017961/1, the Wellcome Innovator Awards 215733/Z/19/Z and 221633/Z/20/Z, CCMI and the Alan Turing Institute. 


\bibliographystyle{splncs04}
\bibliography{augHypergraph}

\begin{thebibliography}{10}
\providecommand{\url}[1]{\texttt{#1}}
\providecommand{\urlprefix}{URL }
\providecommand{\doi}[1]{https://doi.org/#1}

\bibitem{abdar2021review}
Abdar, M., Pourpanah, F., Hussain, S., Rezazadegan, D., Liu, L., Ghavamzadeh, M., Fieguth, P., Cao, X., Khosravi, A., Acharya, U.R., et~al.: A review of uncertainty quantification in deep learning: Techniques, applications and challenges. Information Fusion  \textbf{76},  243--297 (2021)

\bibitem{aviles2022multi}
Aviles-Rivero, A.I., Runkel, C., Papadakis, N., Kourtzi, Z., Sch{\"o}nlieb, C.B.: Multi-modal hypergraph diffusion network with dual prior for alzheimer classification. In: International Conference on Medical Image Computing and Computer-Assisted Intervention. pp. 717--727. Springer (2022)

\bibitem{cubuk2020randaugment}
Cubuk, E.D., Zoph, B., Shlens, J., Le, Q.V.: Randaugment: Practical automated data augmentation with a reduced search space. In: Proceedings of the IEEE/CVF conference on computer vision and pattern recognition workshops. pp. 702--703 (2020)

\bibitem{ding2022data}
Ding, K., Xu, Z., Tong, H., Liu, H.: Data augmentation for deep graph learning: A survey. ACM SIGKDD Explorations Newsletter  \textbf{24}(2),  61--77 (2022)

\bibitem{feng2019hypergraph}
Feng, Y., You, H., Zhang, Z., Ji, R., Gao, Y.: Hypergraph neural networks. In: Proceedings of the AAAI Conference on Artificial Intelligence. vol.~33, pp. 3558--3565 (2019)

\bibitem{gao2022hgnn+}
Gao, Y., Feng, Y., Ji, S., Ji, R.: Hgnn+: General hypergraph neural networks. IEEE Transactions on Pattern Analysis and Machine Intelligence pp. 3181--3199 (2023)

\bibitem{jiang2019dynamic}
Jiang, J., Wei, Y., Feng, Y., Cao, J., Gao, Y.: Dynamic hypergraph neural networks. In: IJCAI. pp. 2635--2641 (2019)

\bibitem{kendall2017uncertainties}
Kendall, A., Gal, Y.: What uncertainties do we need in bayesian deep learning for computer vision? Advances in neural information processing systems  (2017)

\bibitem{liang2023modeling}
Liang, W., Zhang, K., Cao, P., Zhao, P., Liu, X., Yang, J., Zaiane, O.R.: Modeling alzheimers’ disease progression from multi-task and self-supervised learning perspective with brain networks. In: International Conference on Medical Image Computing and Computer-Assisted Intervention (2023)

\bibitem{pan2021characterization}
Pan, J., Lei, B., Shen, Y., Liu, Y., Feng, Z., Wang, S.: Characterization multimodal connectivity of brain network by hypergraph gan for alzheimer’s disease analysis. In: Chinese Conference on Pattern Recognition and Computer Vision (PRCV). pp. 467--478. Springer (2021)

\bibitem{parisot2018disease}
Parisot, S., Ktena, S.I., Ferrante, E., Lee, M., Guerrero, R., Glocker, B., Rueckert, D.: Disease prediction using graph convolutional networks: application to autism spectrum disorder and alzheimer’s disease. Medical image analysis  \textbf{48},  117--130 (2018)

\bibitem{shao2020hypergraph}
Shao, W., Peng, Y., Zu, C., Wang, M., Zhang, D., Initiative, A.D.N., et~al.: Hypergraph based multi-task feature selection for multimodal classification of alzheimer's disease. Computerized Medical Imaging and Graphics  \textbf{80},  101663 (2020)

\bibitem{shao2021hyper}
Shao, W., Xiang, S., Zhang, Z., Huang, K., Zhang, J.: Hyper-graph based sparse canonical correlation analysis for the diagnosis of alzheimer’s disease from multi-dimensional genomic data. Methods  \textbf{189},  86--94 (2021)

\bibitem{shin2020gandalf}
Shin, H.C., Ihsani, A., Xu, Z., Mandava, S., Sreenivas, S.T., Forster, C., Cha, J., Initiative, A.D.N., et~al.: Gandalf: Generative adversarial networks with discriminator-adaptive loss fine-tuning for alzheimer’s disease diagnosis from mri. In: International Conference on Medical Image Computing and Computer-Assisted Intervention. pp. 688--697. Springer (2020)

\bibitem{wang2023hgib}
Wang, S., Aviles-Rivero, A.I., Kourtzi, Z., Sch{\"o}nlieb, C.B.: Hgib: Prognosis for alzheimer's disease via hypergraph information bottleneck. arXiv preprint arXiv:2303.10390  (2023)

\bibitem{yadati2019hypergcn}
Yadati, N., Nimishakavi, M., Yadav, P., Nitin, V., Louis, A., Talukdar, P.: Hypergcn: A new method for training graph convolutional networks on hypergraphs. Advances in neural information processing systems  \textbf{32} (2019)

\bibitem{you2021graph}
You, Y., Chen, T., Shen, Y., Wang, Z.: Graph contrastive learning automated. In: International Conference on Machine Learning. pp. 12121--12132. PMLR (2021)

\bibitem{zhao2022autogda}
Zhao, T., Tang, X., Zhang, D., Jiang, H., Rao, N., Song, Y., Agrawal, P., Subbian, K., Yin, B., Jiang, M.: Autogda: Automated graph data augmentation for node classification. In: Learning on Graphs Conference. pp. 32--1. PMLR (2022)

\bibitem{zhou2021synthesizing}
Zhou, B., Wang, R., Chen, M.K., Mecca, A.P., O’Dell, R.S., Dyck, C.H.V., Carson, R.E., Duncan, J.S., Liu, C.: Synthesizing multi-tracer pet images for alzheimer’s disease patients using a 3d unified anatomy-aware cyclic adversarial network. In: International Conference on Medical Image Computing and Computer-Assisted Intervention. pp. 34--43. Springer (2021)

\bibitem{zhou2006learning}
Zhou, D., Huang, J., Sch{\"o}lkopf, B.: Learning with hypergraphs: Clustering, classification, and embedding. Advances in neural information processing systems  \textbf{19} (2006)

\bibitem{zuo2021multimodal}
Zuo, Q., Lei, B., Shen, Y., Liu, Y., Feng, Z., Wang, S.: Multimodal representations learning and adversarial hypergraph fusion for early alzheimer’s disease prediction. In: Chinese Conference on Pattern Recognition and Computer Vision (PRCV). pp. 479--490. Springer (2021)

\end{thebibliography}

\end{document}